\documentclass{article}
\usepackage{spconf,amsmath,graphicx}
\usepackage[hidelinks]{hyperref}
\usepackage{caption}
\usepackage{subcaption}
\usepackage{booktabs}
\usepackage{multirow}
\usepackage{xcolor}
\usepackage{amsmath}

\title{A Study on the Integration of Pre-trained SSL, ASR, LM and SLU Models for Spoken Language Understanding}

\name{\parbox{\textwidth}{ \centering 
Yifan Peng\sthanks{Equal contribution.}, Siddhant Arora$^{*}$, Yosuke Higuchi, Yushi Ueda, Sujay Kumar, Karthik Ganesan,\\ Siddharth Dalmia, Xuankai Chang, Shinji Watanabe
}}

\address{Carnegie Mellon University, Pittsburgh, PA, USA\\
\texttt{
\footnotesize{\{yifanpen,siddhana,yhiguchi,yueda,sujayk,karthikg,sdalmia,xuankaic,swatanab\}@andrew.cmu.edu}
}}

\copyrightnotice{978-1-6654-7189-3/22/\$31.00~\copyright2023 IEEE}
\begin{document}
\ninept
\maketitle
\begin{abstract}
Collecting sufficient labeled data for spoken language understanding (SLU) is expensive and time-consuming. Recent studies achieved promising results by using pre-trained models in low-resource scenarios. Inspired by this, we aim to ask: which (if any) pre-training strategies can improve performance across SLU benchmarks? To answer this question, we employ four types of pre-trained models and their combinations for SLU. We leverage \textit{self-supervised} speech and language models (LM) pre-trained on large quantities of unpaired data to extract strong speech and text representations. We also explore using \textit{supervised} models pre-trained on larger external automatic speech recognition (ASR) or SLU corpora. We conduct extensive experiments on the SLU Evaluation (SLUE) benchmark and observe \textit{self-supervised} pre-trained models to be more powerful, with pre-trained LM and speech models being most beneficial for the Sentiment Analysis and Named Entity Recognition task, respectively.
\footnote{Our code and models will be publicly available as part of the ESPnet-SLU toolkit.}
\end{abstract}
\section{Introduction}

Spoken language understanding (SLU) aims to extract semantics from a spoken utterance, which is essential for spoken dialog systems, voice assistants and intelligent home devices~\cite{socialbot, snips-voice-platform}. SLU comprises a wide range of tasks, including extracting the intent~\cite{Lugosch_FSC,SLURP,coucke2018snips} and associated entities~\cite{SLURP, earnings21}, recognizing emotion~\cite{iemocap} for a given utterance, or modeling the topic of user conversations~\cite{SWB_DA_res,SWB}. Traditional SLU systems consist of two cascaded modules, i.e., automatic speech recognition (ASR) and natural language understanding (NLU). Recent studies have explored deep learning-based end-to-end (E2E) approaches that directly predict semantic meanings from a speech signal without converting it to intermediate text~\cite{e2e-slu-Haghani, e2e-slu-Serdyuk, e2e-slu-pre-train}. These E2E approaches avoid the error propagation seen in pipeline models as well as can capture non-phonemic signals such as pauses and intonations that a text-based system cannot capture.

However, E2E models usually require a large amount of labeled training data. SLU datasets are often expensive and time-consuming to collect, and hence most publicly available SLU datasets are limited in size. For low-resource applications, researchers have explored pre-trained representations and achieved promising results~\cite{superb, SLUE}. Pre-trained language models like BERT~\cite{BERT} and DeBERTa~\cite{DeBERTa} learn rich textual representations from unlabeled text and are shown to advance the state-of-the-art (SOTA) performance when fine-tuned on downstream NLU tasks. Similarly, self-supervised speech representations can improve various speech processing tasks~\cite{superb, wav2vec2}. 
Inspired by these studies, there has been a lot of interest in pre-training the acoustic~\cite{lai2021semi,e2e-slu-pre-train,pasad2021use} and semantic~\cite{speechbert,agrawal2020tie,chung2020splat,lai2021semi,pasad2021use} model components for SLU tasks on large quantities of unlabeled speech and text data. 

To this end, we ask the following questions: (i) Can pre-training methodologies help to advance performance across various SLU benchmarks? (ii) Which pre-training methodologies are most useful to improve performance for a given SLU task? (iii) Can we identify the kind of spoken utterances that are responsible for the majority of performance gains achieved by a given pre-training strategy? We seek to answer these questions by conducting a thorough study of various pre-training paradigms in the context of SLU. We investigate the following four types of pre-trained models and their combinations: 1) \textit{self-supervised} learning (SSL) speech models~\cite{tera,vq-apc,wav2vec2,hubert,wavlm}, to generate powerful acoustic representations from the raw audio; 2) \textit{self-supervised} language models (LM)~\cite{BERT,DeBERTa}, to build strong semantic representations; 3) \textit{supervised} ASR models pre-trained on large corpora~\cite{gigaspeech,spgispeech} and 4) \textit{supervised} SLU models pre-trained on other SLU corpora~\cite{iemocap,swbd_sentiment}. We conduct extensive experiments on the newly released Spoken Language Understanding Evaluation (SLUE) benchmark~\cite{SLUE}, which provides well-designed datasets with baselines and metrics for evaluating low-resource SLU. It consists of two SLU tasks: sentiment analysis (SA) on SLUE-VoxCeleb and named entity recognition (NER) on SLUE-VoxPopuli, with a small amount of labeled data to fine-tune the SLU system. This makes it an interesting benchmark to evaluate the efficacy of different pre-training approaches since the SLU dataset is particularly under-resourced. 

Our contributions are as follows:
\begin{itemize}
    \item We investigate the efficacy of four types of pre-trained models and their integrations in the context of SLU.
    \item We conduct extensive experiments on the SLUE benchmark and show that each pre-training approach can improve performance over the baseline E2E model without pre-training. Our best models can outperform the baseline by a large margin on both SA and NER tasks.
    \item Our results demonstrate that pre-training methodologies based on \textit{self-supervised} learning are more powerful than those based on \textit{supervised} learning. We hypothesize that this is because \textit{self-supervised} models are trained on huge amounts of unlabeled data, which have extensive coverage of acoustic and linguistic variations. We also observe strong semantic representations from a pre-trained LM DeBERTa to be most helpful for the SA task, whereas strong speech representations produced by an SSL model WavLM to be most beneficial for the NER task. We believe that future research to build SLU systems should employ pre-training paradigms based on \textit{self-supervised} representations to boost model performance, particularly in low resource scenarios.
    \item We analyze the performance gains from pre-training techniques and find that most of the performance improvement from \textit{self-supervised} pre-training methods can be seen in semantically and acoustically challenging utterances.
    \item Another interesting finding from our experiments is that the word error rate (WER) in ASR transcripts is not very well correlated with the downstream SA task but is a good indicator of the downstream NER performance. 
\end{itemize}

\section{Methods}

\subsection{Problem formulation}
\label{sec:formulation}
\begin{figure}[t]
\centering
\begin{subfigure}[b]{\linewidth}
\centering
\includegraphics[width=\textwidth]{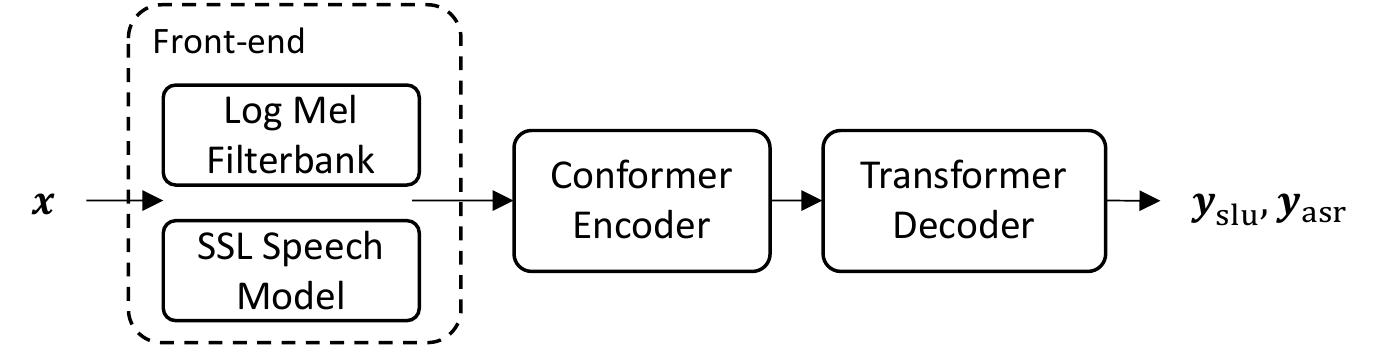}
\caption{Our E2E SLU system. Self-supervised speech representations can replace the log Mel filterbank features. The entire model can be pre-trained on external ASR or SLU corpora.}
\label{fig:e2e}
\end{subfigure}
\begin{subfigure}[b]{\linewidth}
\centering
\includegraphics[width=\textwidth]{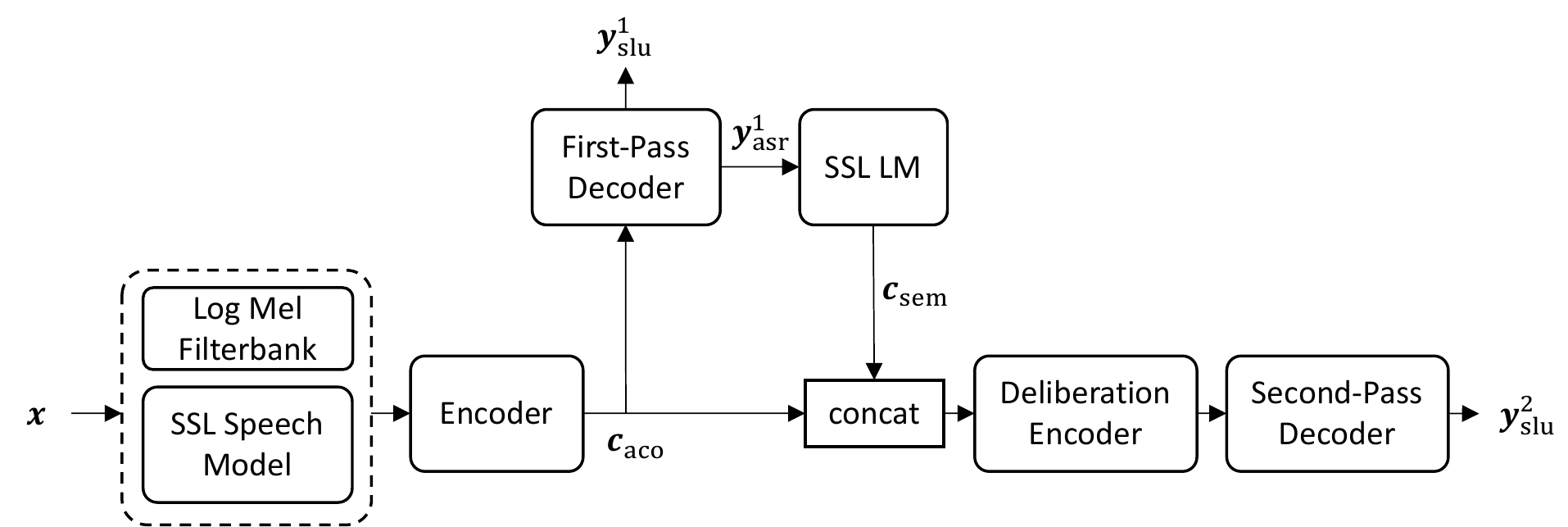}
\caption{Our two-pass SLU system~\cite{2-pass-slu}. Self-supervised speech representations can replace the log Mel filterbank features. The $2^{nd}$ pass attends to both acoustic information from $1^{st}$ pass and semantic information from ASR transcript, as discussed in Section~\ref{sec:ssl-lm}.}
\label{img:Semantic_SLU}
\end{subfigure}
\caption{Overview of our SLU systems.}
\label{fig:system-overview}
\vskip -0.15in
\end{figure}

As in ESPnet-SLU~\cite{espnet-slu}, we formulate the two SLU tasks (i.e., SA and NER) as a unified sequence-to-sequence problem. The input is a sequence of speech features extracted from the raw audio, and the output is a sequence of tokens consisting of the transcript and SLU labels. For SA, a sentiment label is prepended to the transcript. For NER, each entity phrase in the transcript begins with an entity tag and ends with a special token, which is consistent with the SLUE toolkit~\cite{SLUE}.
Figure~\ref{fig:system-overview} shows our SLU systems. The attention-based encoder-decoder architecture is adopted in our end-to-end (E2E) approaches. Specifically, we employ the Conformer~\cite{conformer} encoder and Transformer~\cite{transformer} decoder. Note that we do not use a language model for decoding. To better incorporate semantic information, we also exploit a two-pass approach~\cite{2-pass-slu}, as introduced in Section~\ref{sec:ssl-lm}.

\subsection{Self-supervised pre-trained speech models}
\label{subsec:ssl}
Self-supervised speech models are pre-trained on large volumes of unlabeled speech data and can generate powerful representations for downstream tasks, especially for low-resource applications. We employ pre-trained speech representations to replace the commonly used log Mel filterbank features. Following prior work~\cite{superb, espnet-ssl}, a weighted sum of multiple hidden states is utilized. During training, the parameters of pre-trained models are frozen and never updated. Five self-supervised speech models are evaluated, namely TERA~\cite{tera}, VQ-APC~\cite{vq-apc}, Wav2Vec2~\cite{wav2vec2}, HuBERT~\cite{hubert} and WavLM~\cite{wavlm}. These models are trained using different objectives and corpora (see Table~\ref{tab:ssl-speech}).

\subsection{Self-supervised pre-trained language models}
\label{sec:ssl-lm}
Self-supervised LMs are pre-trained on a large amount of unlabeled text data, which can generate high-quality semantic representations. There are multiple ways to utilize pre-trained LMs, such as pipeline approaches~\cite{SLUE} and jointly modeling speech and text in a shared latent space~\cite{chung2020splat}. Inspired by prior work on two-pass ASR~\cite{hu2020deliberation,sainath2019two}, we adopt a two-pass SLU approach \cite{2-pass-slu} in this work, where the first pass predicts SLU labels and ASR hypotheses from the audio, and the second pass improves on the initial prediction by combining both acoustic and semantic information from ASR hypotheses. More specifically, the input speech ($x$) is first passed through an acoustic encoder to generate acoustic embeddings ($c_{\text{aco}}$). These embeddings are then passed to the first-pass decoder, which predicts the first-pass SLU labels ($y^{1}_{\text{slu}}$) and ASR transcript ($y^{1}_{\text{asr}}$). 

The ASR transcript is then tokenized and processed by a pre-trained LM to generate semantic embeddings ($c_{\text{sem}}$). The acoustic and semantic embeddings are concatenated along the time dimension to form a joint embedding. Finally, the joint embedding is passed through a deliberation encoder before entering the second-pass decoder to predict more accurate second-pass SLU labels ($y^{2}_{\text{slu}}$). In this work, we have only experimented with BERT~\cite{BERT} and DeBERTa~\cite{DeBERTa}; however, our method can incorporate any of the pre-trained models provided by HuggingFace.~\footnote{\url{https://huggingface.co/models}}

\subsection{Supervised pre-trained ASR models}
\label{ssec:asr}
To compensate for the lack of labeled data in the under-resourced SLU task, we explore pre-training an SLU model using a large-scale external ASR corpus. This pre-training is expected to initialize the model with strong acoustic processing ability, which can improve performance on the downstream SLU task when fine-tuned on the target dataset.

In this work, we adopt GigaSpeech~\cite{gigaspeech} and SPGISpeech~\cite{spgispeech} for the ASR pre-training, which are publicly available corpora consisting of 10k and 5k hours of transcribed English speech, respectively. We then initialize the SLU model with the pre-trained parameters, except for the embedding and softmax layers in the encoder and decoder networks.
Then, the model is fine-tuned for a target task using a small amount of labeled data.

\begin{table}[t]
  \caption{Summary of self-supervised pre-trained speech models used in this work. The Mix 94k dataset is a mixture of LibriLight 60k~\cite{librilight}, GigaSpeech 10k~\cite{gigaspeech}, and VoxPopuli 24k~\cite{voxpopuli}.}
  \label{tab:ssl-speech}
  \centering
  \resizebox{\linewidth}{!}{
  \begin{tabular}{cccc}
    \toprule
    Model & Architecture & Dataset & Objective\\
    \midrule
    TERA & 3-Trans & LibriSpeech 960h & masking\\
    VQ-APC & 3-GRU & LibriSpeech 960h & auto-regressive\\
    Wav2Vec2 & 7-Conv 24-Trans & LibriLight 60k & contrastive\\
    HuBERT & 7-Conv 24-Trans & LibriLight 60k & pseudo-labeling\\
    WavLM & 7-Conv 24-Trans & Mix 94k & pseudo-labeling\\
    \bottomrule
  \end{tabular}
  }
\end{table}

\begin{table}[t]
  \caption{Overview of the two datasets in the SLUE benchmark~\cite{SLUE}.}
  \label{tab:slue-datasets}
  \centering
  \resizebox{\linewidth}{!}{
  \begin{tabular}{llccc}
    \toprule
    \multirow{2}{*}{Dataset} & \multirow{2}{*}{Tasks} & \multicolumn{3}{c}{Size (utterances / hours)}\\\cmidrule(lr){3-5}
    & & Train & Dev & Test\\
    \midrule
    SLUE-VoxCeleb & ASR, SA & 5,777 / 12.8 & 955 / 2.1 & 4,052 / 9.0\\
    SLUE-VoxPopuli & ASR, NER & 5,000 / 14.5 & 1,753 / 5.0 & 1,842 / 4.9\\
    \bottomrule
  \end{tabular}
  }
\vskip -0.1in
\end{table}

\subsection{Supervised pre-trained SLU models}
When the target task has limited data, it is natural to pre-train the SLU model using other corpora designed for a similar task and then fine-tune it for the target task.
For SA, we use existing emotion or sentiment datasets, IEMOCAP~\cite{iemocap} and Switchboard (SWBD) Sentiment~\cite{swbd_sentiment}, which contain 12 and 140 hours of labeled speech data, respectively.
The nine emotion labels in IEMOCAP can be optionally converted into three sentiment labels in the following manner: \{happiness, excited, surprised: Positive\}, \{neutral: Neutral\}, and \{fear, sadness, anger, frustration, disgust: Negative\}. The SWBD Sentiment dataset has three sentiment labels (Positive, Neutral, Negative). Thus, the labels in the two datasets can be preprocessed to match the three labels used in the SLUE SA task.
For NER, we pre-train the model on the SLURP~\cite{SLURP} dataset, which contains about 100 hours of audio data collected from single-turn user interactions with a home assistant. We use the original entity tags for pre-training.

\vskip -0.1in
\subsection{Combination of pre-trained models}
\label{ssec:comb}
The pre-trained models can be combined at either model-level or output-level. For model-level combination, we employ SSL speech representations to replace the log Mel filterbank features. We then pre-train the entire encoder-decoder model on external corpora and fine-tune it using the target SLU dataset. Besides, we combine the self-supervised speech and language models in our two-pass approach (Section~\ref{sec:ssl-lm}).
For output-level combination, we adopt voting-based strategies to aggregate the decoded sequences from different models. We apply the majority voting to obtain SA results. As introduced in Section~\ref{sec:formulation}, the named entity tags are inserted into the transcript, so we directly apply the recognizer output voting error reduction (ROVER)~\cite{ROVER} method to combine multiple hypotheses and extract NER results from the combined word sequence. We also obtain the ASR results for both tasks using ROVER.

\section{Experimental setup}

\subsection{Datasets and tasks}
We adopt the recently released SLUE benchmark~\cite{SLUE} for evaluation, which focuses on naturally produced speech and contains limited labeled training data. The SLUE benchmark consists of two well-designed datasets, i.e., SLUE-VoxCeleb and SLUE-VoxPopuli, and three tasks, i.e., SA, NER and ASR. Specifically, SLUE-VoxCeleb is used for ASR and SA, while SLUE-VoxPopuli is used for ASR and NER. Details about each dataset are shown in Table~\ref{tab:slue-datasets}. The training set is very limited, which is suitable for evaluating low-resource SLU. The released test sets are blind without groundtruth labels. We compare different methods using the development set.

\subsection{Evaluation metrics}
\label{sec:metrics}
We adopt the evaluation metrics in the SLUE benchmark~\cite{SLUE}. ASR is evaluated using word error rate (WER). SA aims to classify an input utterance as having negative, neutral, or positive sentiment, which is evaluated using macro-averaged (unweighted) recall and F1 scores. Since the negative class has only 3 instances in the official development set, we find the results to be unstable.~\footnote{A small change in the negative class will drastically affect the overall metric. For example, if a model achieves a recall of 60\% in both positive and neutral classes but fails to recognize any negative instances, the macro-averaged recall will be 40\%. If it happens to recognize one of the three negative instances, the recall of the negative class will be 33\% and the macro-recall will be 51\%. This means the metric averaged over all three classes is very unstable.} Hence, we use the macro F1 (F1*) and recall (Recall*) computed only on positive and neutral classes for model comparison.~\footnote{After detailed discussions with the authors of the SLUE benchmark.} NER focuses on recognizing the named entities and their tags (types) in a given utterance. There are 7 distinct entity tags. We use micro-averaged F1 and label-F1 scores for NER. The F1 score considers an unordered list of named entity phrases and tag pairs, while the label-F1 only considers the tag predictions and ignores the potential misspelling and segmentation errors in speech recognition.

\subsection{Implementation Details}

Our models are implemented in PyTorch~\cite{pytorch}, and the experiments are conducted using the ESPnet-SLU toolkit~\cite{espnet-slu, espnet}. The self-supervised pre-trained speech and language models are obtained from S3PRL~\cite{superb}, Fairseq~\cite{fairseq} and HuggingFace~\cite{wolf-etal-2020-transformers}, while the pre-trained ASR and SLU models are downloaded from ESPnet Model Zoo. Our models are based on the attention-based encoder-decoder architecture described in Section~\ref{sec:formulation}. 
The encoder is a 12-layer Conformer~\cite{conformer}, while the decoder is a 6-layer Transformer~\cite{transformer}. The number of heads and dimension of a self-attention layer are set to 4 and 256, respectively. The linear units are 1024 for the encoder and 2048 for the decoder. During training, speed perturbation and SpecAugment~\cite{specaugment} are performed for data augmentation. 
We apply dropout~\cite{dropout} and label smoothing~\cite{label-smoothing} to mitigate overfitting. We use the Adam~\cite{adam} optimizer with a maximum learning rate of 2e-3 and a weight decay of 1e-6 or 1e-5. We also employ the Transformer learning rate scheduler~\cite{transformer} with 5k warmup steps. 
We perform joint CTC-attention training and decoding~\cite{joint-ctc-att-mtl, joint-ctc-att-decoding}. 
More details about our models and the config files will be publicly available  as part of the ESPnet-SLU~\cite{espnet-slu} toolkit.

\section{Results}
Table~\ref{tbl:voxceleb} presents the SA results on SLUE-VoxCeleb, and Table~\ref{tbl:voxpopuli} shows the NER results on SLUE-VoxPopuli. In the following subsections, we discuss the effect of different pre-trained models on the performance of these SLU tasks.

\subsection{Sentiment Analysis}
\label{ssec:result_sa}
Table~\ref{tbl:voxceleb} shows that all pre-trained approaches boost SA performance on the SLUE-VoxCeleb dataset.
Among all the models with SSL features, the model using WavLM features achieves the best recall (Recall*: 66.9) and F1 (F1*: 66.9).
Further, models with HuBERT and Wav2Vec2 also outperform the baseline, showing that self-supervised speech representations are more powerful than the log Mel features in the low-resource scenario.

For models pre-trained on large external ASR data, pre-training on GigaSpeech (Recall*: 66.3, F1*: 66.6) achieves better scores than pre-training on SPGISpeech, because GigaSpeech has larger and more diverse training data. However, the best \textit{supervised} model pre-trained on external ASR data has lower performance than the best model using \textit{self-supervised} speech representations.

\begin{table}[t]
\caption{Macro-averaged recall (Recall*) and F1 (F1*) scores (\%) computed only on positive and neutral classes for sentiment analysis on SLUE-VoxCeleb. As discussed in section~\ref{sec:metrics}, we use Recall* and F1* for model comparison. LM decoding is not used. Bold values indicate 
the best performance obtained both with and without output-level system combinations on this dataset and $\underbar{\text{X}}$ indicates outperforming the SA performance of no pre-train model.}
\label{tbl:voxceleb}
\centering
\resizebox {\linewidth} {!} {
\begin{tabular}{ll|ccc}
\toprule
& Pre-trained Model/Corpus & Recall* ($\uparrow$) & F1* ($\uparrow$) & WER ($\downarrow$) \\ \midrule
\multicolumn{3}{l}{\textbf{Our E2E approaches}} \\
\quad w/o pre-train & N/A & 62.4 & 63.6 & 33.0\\\midrule
\quad \multirow{5}{*}{w/ SSL} 
& TERA & $\underbar{\text{62.5}}$ & 62.4 & 27.1 \\ 
& VQ-APC & 61.3 & 62.1 & 29.8  \\ 
& Wav2Vec2  & $\underbar{\text{64.5}}$ & $\underbar{\text{64.4}}$ & 14.2\\
& HuBERT & $\underbar{\text{64.5}}$  & $\underbar{\text{65.2}}$ & 12.8 \\ 
& WavLM & $\underbar{\text{\textbf{66.9}}}$ & $\underbar{\text{66.9}}$ & \hphantom{0}9.1\\ \midrule
\quad \multirow{2}{*}{w/ ASR} 
& GigaSpeech & $\underbar{\text{66.3}}$ & $\underbar{\text{66.6}}$ & 11.3\\
& SPGISpeech & $\underbar{\text{63.3}}$ & $\underbar{\text{64.1}}$ & 14.2\\ \midrule
\quad \multirow{2}{*}{w/ SLU}
& IEMOCAP & 62.4 & 62.9 & 33.2 \\
& SWBD Sentiment & $\underbar{\text{64.7}}$ & $\underbar{\text{64.8}}$ & 21.9\\
& WavLM+IEMOCAP & $\underbar{\text{64.3}}$ & $\underbar{\text{65.2}}$ & \hphantom{0}9.3\\
& WavLM+SWBD Sentiment & $\underbar{\text{64.9}}$ & $\underbar{\text{65.7}}$ & \hphantom{0}9.0 \\ \midrule
\quad \multirow{2}{*}{w/ LM (2-pass)}
& BERT & $\underbar{\text{64.7}}$ & $\underbar{\text{65.1}}$ & 29.5\\
& DeBERTa & $\underbar{\text{66.2}}$ & $\underbar{\text{\textbf{67.3}}}$ & 29.5\\
& WavLM+BERT &  $\underbar{\text{66.8}}$ & $\underbar{\text{65.7}}$ & \hphantom{0}9.4\\
& WavLM+DeBERTa & $\underbar{\text{\textbf{66.9}}}$ & $\underbar{\text{66.5}}$ & \hphantom{0}9.3 \\
\midrule
\midrule
\multicolumn{3}{l}{\textbf{Our system combination}}\\
\quad majority voting & best 3 models & $\underbar{\text{\textbf{67.9}}}$ & $\underbar{\text{\textbf{69.0}}}$ & 9.7\\
\bottomrule
\end{tabular}
}
\vskip -0.15in
\end{table}

\begin{table}[t]
\caption{Macro-averaged recall and F1 scores (\%) for sentiment analysis on SLUE-VoxCeleb. As discussed in Section~\ref{sec:metrics}, we actually do not use these Recall and F1 for model comparison since they are unstable, but we do show that we outperform the results reported in the SLUE benchmark using a similar Wav2Vec2 based SLU model.}
\label{tbl:voxceleb_small}
\centering
\resizebox {\linewidth} {!} {
\begin{tabular}{ll|ccc}
\toprule
& Pre-trained Model & Recall ($\uparrow$) & F1 ($\uparrow$) & WER ($\downarrow$) \\ \midrule
\multicolumn{5}{l}{\textbf{SLUE benchmark}~\cite{SLUE}} \\
\quad \multirow{2}{*}{Oracle Text} & BERT & 43.0 & 43.6 & 0.0\\
& DeBERTa & 55.6 & 46.5 & 0.0\\\midrule
\quad Pipeline w/o LM & Wav2Vec2+DeBERTa & 54.2 & 45.3 & 11.0\\
\quad Pipeline w/ LM & Wav2Vec2+DeBERTa & 55.1 & 45.8 & \hphantom{0}9.1 \\\midrule
\quad E2E & Wav2Vec2 w/o LM & 45.0 & 44.2 & 11.0\\
\quad E2E & Wav2Vec2 w/ LM & 45.0 & 44.2 & \hphantom{0}9.1\\\midrule
\midrule
\multicolumn{3}{l}{\textbf{Our E2E approach}} \\
\quad w/ SSL 
& Wav2Vec2  & 54.1 & 46.4 & 14.2\\
\bottomrule
\end{tabular}
}
\vskip -0.1in
\end{table}

For models pre-trained on external sentiment datasets, the performance is higher when using the SWDB dataset, possibly due to the larger size of SWDB sentiment compared to IEMOCAP. 
Two-pass SLU models (Section~\ref{sec:ssl-lm}) that use LMs to enhance semantic processing power perform better than those pre-trained on external SLU datasets. By incorporating DeBERTa as a pre-trained LM (F1*:67.3), we can outperform, in terms of F1 value, all other pre-training paradigms for SA.

Our results demonstrate that \textit{self-supervised} speech models and LMs can generate more robust speech and semantic representations, respectively, underscoring the efficacy of leveraging \textit{self-supervised} models to pre-train SLU systems.
Another remarkable observation is that the WER of the two-pass SLU model with DeBERTa (WER: 29.5, F1*: 67.3) is much higher than that of the model using a WavLM frontend (WER: 9.1, F1*: 66.9) but the two-pass approach still achieves better SA performance. This result shows that WER in ASR transcripts is not a good indicator of the downstream SA performance.

To compare with the E2E results from the SLUE~\cite{SLUE} benchmark, we also report the original Recall and F1 values in Table~\ref{tbl:voxceleb_small}. Our approach shows higher performance using the same Wav2Vec2 model (54.1 vs.\ 45.0 in recall, 46.4 vs.\ 44.2 in F1), demonstrating the effectiveness of our encoder-decoder-based SLU modeling.

We further investigate integrating speech and text pre-training approaches at the model-level using WavLM features as input for our two-pass SLU model (see WavLM+BERT, WavLM+DeBERTa in Table~\ref{tbl:voxceleb}) as well as pre-training on external SLU datasets (see WavLM+IEMOCAP, WavLM+SWDB Sentiment in Table~\ref{tbl:voxceleb}). We generally observe an improvement in performance in comparison to the model that uses log Mel Filterbank features, except for the two-pass SLU model with DeBERTa. We further experiment with output-level combinations (see Section~\ref{ssec:comb}) of our best three models, i.e., pre-trained with WavLM, GigaSpeech, and DeBERTa. As shown in Table~\ref{tbl:voxceleb} (see ``Our system combinations''), this output-level model combination (Recall*: 67.9, F1*: 69.0) outperforms all the individual models, indicating that we can advance the performance on SLU benchmarks by combining different pre-training approaches.

\begin{table}[t]
\caption{Micro-averaged F1 and label-F1 scores (\%) for named entity recognition on SLUE-VoxPopuli. LM decoding is not used in our approaches. Bold values indicate 
the best performance obtained both with and without output-level system combinations on this dataset and $\underbar{\text{X}}$ indicates outperforming the NER performance of no pre-train model.}
\label{tbl:voxpopuli}
\centering
\resizebox {\linewidth} {!} {
\begin{tabular}{l @{\hspace{1\tabcolsep}} l @{\hspace{0.5\tabcolsep}} | @{\hspace{0.5\tabcolsep}} c @{\hspace{1\tabcolsep}} c @{\hspace{1\tabcolsep}} c}
\toprule
& Pre-trained Model/Corpus  & Label-F1 ($\uparrow$) & F1 ($\uparrow$) & WER ($\downarrow$) \\ \midrule
\multicolumn{5}{l}{\textbf{SLUE benchmark}~\cite{SLUE}}\\
\quad \multirow{2}{*}{Oracle Text} 
& BERT & 90.9 & 86.2 & \hphantom{0}0.0\\
& DeBERTa & 91.1 & 87.5 & \hphantom{0}0.0\\\midrule
\quad Pipeline w/o LM & Wav2Vec2+DeBERTa & 83.5 & 63.3 & 14.0\\
\quad Pipeline w/ LM & Wav2Vec2+DeBERTa & 87.4 & 74.9 & \hphantom{0}9.1\\\midrule
\quad E2E w/o LM & Wav2Vec2 & 69.1 & 55.6 & 14.0\\
\quad E2E w/ LM & Wav2Vec2 & 79.0 & 70.2 & \hphantom{0}9.1\\
\midrule\midrule
\multicolumn{5}{l}{\textbf{Our E2E approaches}}\\
\quad w/o pre-train & N/A & 67.6 & 54.7 & 34.2\\\midrule
\quad \multirow{5}{*}{w/ SSL}
& TERA & $\underbar{\text{70.9}}$  & $\underbar{\text{57.1}}$ & 28.6\\ 
& VQ-APC & $\underbar{\text{76.6}}$  & $\underbar{\text{63.6}}$ & 24.1\\ 
& Wav2Vec2 & $\underbar{\text{83.3}}$ & $\underbar{\text{69.5}}$ & 12.8\\
& HuBERT & $\underbar{\text{84.8}}$  & $\underbar{\text{69.7}}$ & 12.5\\ 
& WavLM & $\underbar{\text{\textbf{88.0}}}$  & $\underbar{\text{74.5}}$ & \hphantom{0}9.3\\ \midrule
\quad \multirow{2}{*}{w/ ASR}
& GigaSpeech & $\underbar{\text{86.0}}$ & $\underbar{\text{73.9}}$ &  11.2 \\
& SPGISpeech & $\underbar{\text{84.1}}$ & $\underbar{\text{71.4}}$ & 12.2 \\ \midrule
\quad w/ SLU %
& SLURP & $\underbar{\text{71.5}}$ & $\underbar{\text{59.7}}$ & 33.7 \\
& WavLM+SLURP & $\underbar{\text{87.5}}$ & $\underbar{\text{\textbf{75.7}}}$ & \hphantom{0}9.0\\\midrule
\quad \multirow{2}{*}{w/ LM (2-pass)}
& BERT & $\underbar{\text{69.2}}$ & 54.5 & 33.8\\
& DeBERTa & $\underbar{\text{69.4}}$ & $\underbar{\text{55.5}}$ & 34.1\\
& WavLM+BERT & $\underbar{\text{87.3}}$ & $\underbar{\text{73.5}}$ & \hphantom{0}9.6\\
& WavLM+DeBERTa & $\underbar{\text{87.7}}$ & $\underbar{\text{74.0}}$ & \hphantom{0}9.5\\\midrule
\midrule
\multicolumn{5}{l}{\textbf{Our system combination}}\\
\quad ROVER & best 4 models & $\underbar{\text{\textbf{88.7}}}$ & $\underbar{\text{\textbf{77.2}}}$ & \hphantom{0}8.6\\
\bottomrule
\end{tabular}
}
\vskip -0.2in
\end{table}

\subsection{Named Entity Recognition}
Table~\ref{tbl:voxpopuli} shows that similar to SA, all pre-training approaches improve NER performance over the baseline SLU model without pre-training. WavLM (F1: 74.5, Label F1: 88.0) performs the best in all SSL models, and our findings for the utility of different pre-trained SSL systems as feature extractors are mainly consistent with the SA task. We also observe that pre-training on external ASR data, particularly the GigaSpeech dataset (F1: 73.9, Label F1: 86.0), boosts performance but is still worse than using WavLM features. 

We observe that models incorporating strong speech representations from the SSL model or external ASR dataset generally have better performance than those using semantic representations obtained through pre-trained LM or external SLU corpora. Using the \textit{self-supervised} WavLM model to extract speech features is found to be most beneficial for advancing the NER performance of SLU systems. We also find the NER performance of all encoder-decoder models to be well correlated with the WER of ASR transcripts.

We compare our models with the results from the SLUE~\cite{SLUE} benchmark and observe that even without LM decoding, our best model using WavLM features outperforms the SLUE E2E approach with LM decoding (88.0 vs.\ 79.0 in label-F1, 74.5 vs.\ 70.2 in F1).

We also experiment with a model that takes WavLM features as input and pre-trains using external LMs or SLU corpora (see WavLM+BERT, WavLM+DeBERTa, WavLM+SLURP in Table~\ref{tbl:voxpopuli}) and conclude that using WavLM features inside this pre-training framework can boost the NER performance. Further, by pre-training on the SLURP dataset, the model that uses WavLM features can achieve an F1 score of 75.7, which is higher than the model pre-trained using only WavLM features (F1: 74.5).
This result shows that gains achieved by pre-training on SLURP are complementary to strong speech representations obtained from WavLM and provides evidence that integration of different pre-training paradigms at the model-level can advance the NER performance. We then investigate the output-level combination (see Section~\ref{ssec:comb}) of our best four models, i.e., pre-trained with WavLM, model-level combination of WavLM and SLURP, model-level combination of WavLM and DeBERTa and GigaSpeech. Results show that integrating different pre-training approaches using ROVER (see ``Our system combinations'' in Table~\ref{tbl:voxpopuli}) can further boost NER performance, which encourages future research on combinations of different pre-training approaches.

\vskip -1.0in
\section{Analysis}
\vskip -0.1in
Recently, there has been interest~\cite{semcomplex2020} in quantifying the semantic complexity of SLU datasets and reporting the performance of a given SLU system across different semantic complexities. Prior work~\cite{MASE_Eval} has also shown that ASR pre-training can be particularly useful for acoustically and semantically challenging utterances. Inspired by these findings, we also compare the performance of our trained models across utterances of different acoustic and semantic complexities. To facilitate this analysis, we divide the development set into classes of different difficulties with roughly similar number of utterances. Our analysis helps us develop a deep understanding of gains achieved by best performing pre-trained models on each of the SLU tasks, i.e., using \textit{self-supervised} DeBERTa for the SA task and \textit{self-supervised} WavLM for the NER task. We further compare performance between \textit{supervised} and \textit{self-supervised} pre-training methodologies and characterize the utterances responsible for the performance gap between the two approaches. %

\subsection{Acoustic Analysis}

\begin{figure}[t]
\centering
\includegraphics[width=0.9\linewidth]{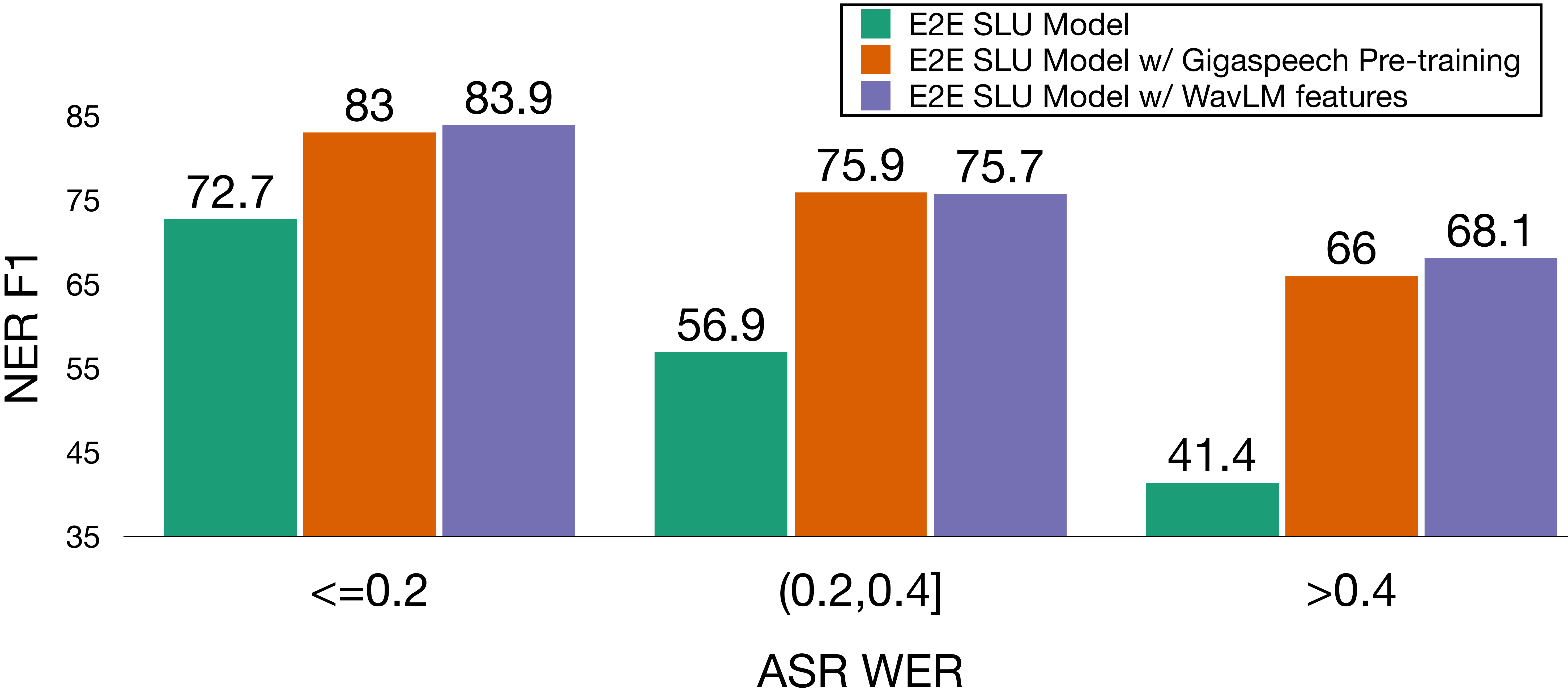}
\caption{Results comparing the NER performance of our models without pre-training, with ASR pre-training on GigaSpeech and with WavLM features as input across different ASR difficulties measured by WER of the no pre-train model on SLUE-VoxPopuli. The performance gain from using \textit{self-supervised} WavLM increases as the ASR difficulty increases.}
\label{fig:wer_breakdown}
\vskip -0.1in
\end{figure}
\begin{figure}[t]
\centering
\includegraphics[width=0.9\linewidth]{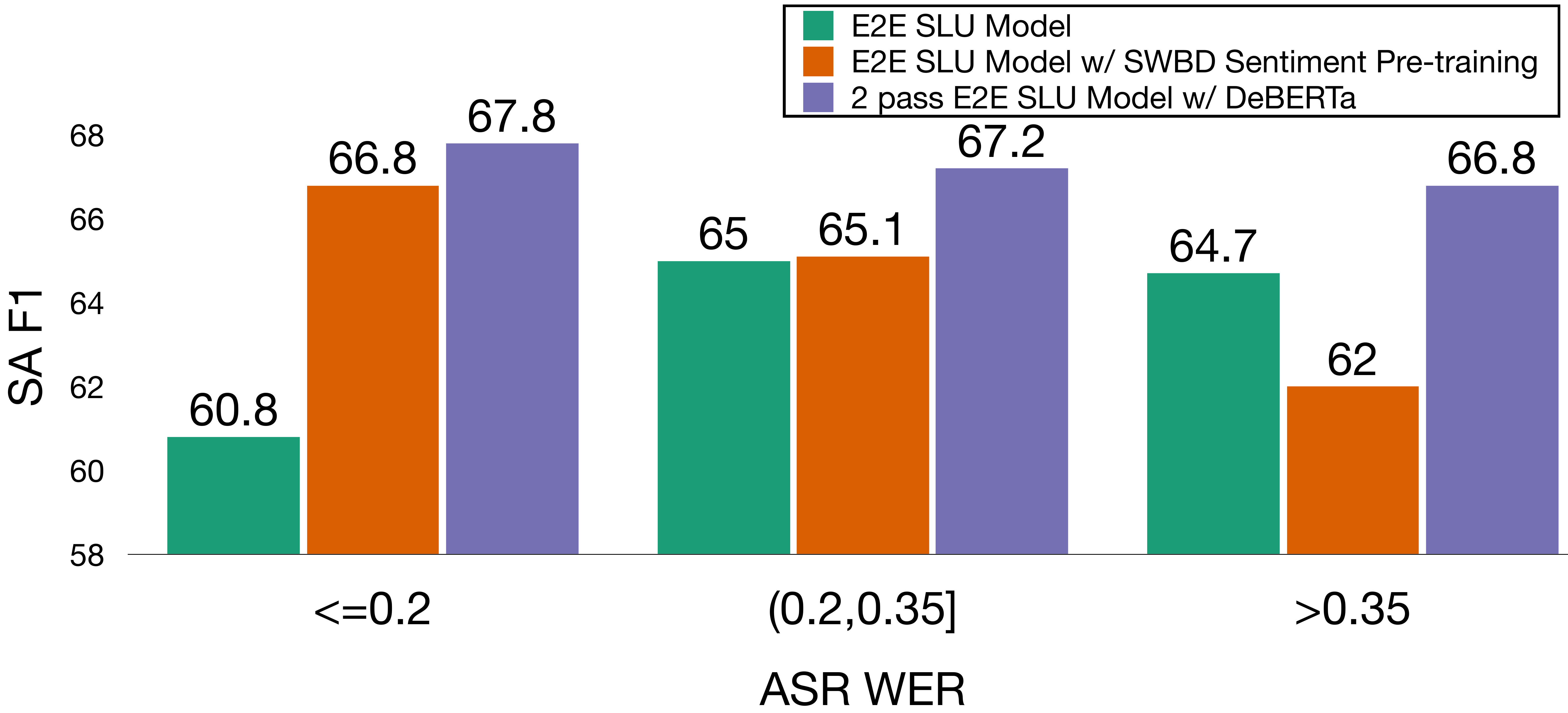}
\caption{Results comparing the SA performance of our models without pre-training, with SLU pre-training on SWDB Sentiment and with DeBERTa as a pre-trained LM across different ASR difficulties measured using WER of the no pre-train model on SLUE-VoxCeleb. The largest performance gap between the model using \textit{self-supervised} DeBERTA and the model pre-trained on SWDB Sentiment is from acoustically challenging utterances (WER$>0.35$).}
\label{fig:sa_wer_breakdown}
\vskip -0.25in
\end{figure}

Figure~\ref{fig:wer_breakdown} analyzes the gains in NER performance by using SSL features as input. 
We quantify the acoustic complexity of spoken utterance using WER of ASR transcripts produced by our baseline E2E SLU model.
We observe that the performance gap between the baseline model and the pre-trained models increases as the ASR difficulty of utterances increases in the VoxPopuli dataset. 
Further, when we compare using \textit{self-supervised} WavLM features and \textit{supervised} pre-training on GigaSpeech, most of the performance difference is observed in extremely difficult utterances (i.e., WER $>0.4$). Hence, we infer that SSL features are particularly beneficial for acoustically challenging utterances.

Prior work~\cite{2-pass-slu} has shown that better semantic modeling can help recover from ASR errors and hence can improve performance on acoustically challenging utterances. Inspired by this, we similarly analyze the performance gains of the two-pass SLU model with DeBERTa across different acoustic complexities in Figure~\ref{fig:sa_wer_breakdown}. We do not observe any clear trend between SA performance and ASR WER for our models on the VoxCeleb dataset, which is consistent with our findings in Section~\ref{ssec:result_sa}. Interestingly, we observe that acoustically difficult utterances (WER $>0.35$) account for the maximum performance gap between models pre-trained with \textit{self-supervised} DeBERTa and \textit{supervised} SWBD Sentiment SLU model, demonstrating that LM features are more robust to errors in ASR transcript.

\subsection{Semantic Analysis}
\begin{figure}[t]
\centering
\begin{subfigure}[b]{\linewidth}
\centering
\includegraphics[width=0.9\textwidth]{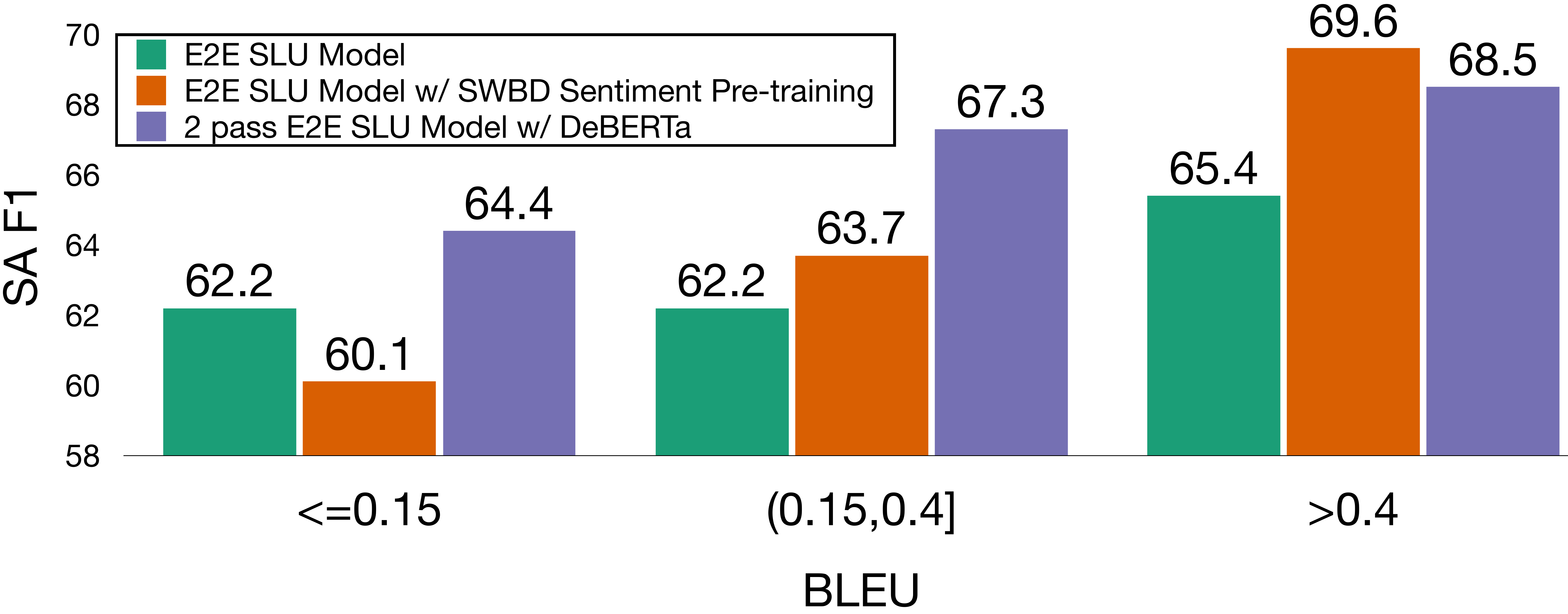}
\caption{SA performance across different lexical overlap (measured by BLEU score) of test utterances with the training set.}
\label{fig:bleu_breakdown_lm}
\end{subfigure}
\begin{subfigure}[b]{\linewidth}
\centering
\includegraphics[width=0.9\linewidth]{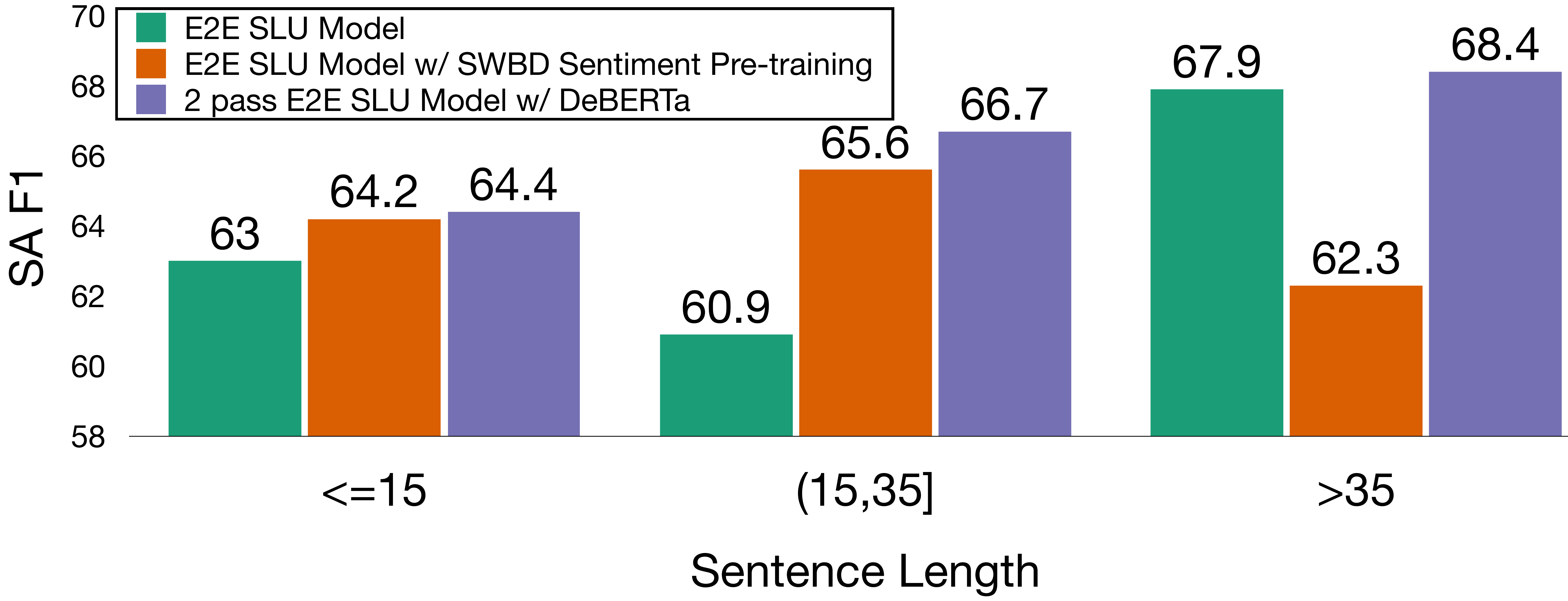}
\caption{SA performance across length of test utterances.}
\label{fig:sen_len_breakdown_lm}
\end{subfigure}
\caption{Results comparing the SA performance of models without pre-training, with SLU pre-training on SWDB Sentiment and with DeBERTa as a pre-trained LM across different semantic difficulties on SLUE-VoxCeleb. The semantic difficulty is measured using (a) lexical overlap with training utterances and (b) utterance length. We observe that \textit{self-supervised} DeBERTa representations are particularly useful for semantically complex utterances which have low n-gram overlap with the training set or longer utterance length.}
\vskip -0.2in
\label{fig:lm-breakdown}
\end{figure}
Figure~\ref{fig:lm-breakdown} analyzes the performance of the two-pass SLU model that uses DeBERTa as a pre-trained LM. A spoken utterance with many unique n-grams not seen in training utterances makes the semantic understanding task more challenging~\cite{MASE_Eval}. As a result, we categorize the test utterances based on their n-gram overlap with training utterances. We choose the Sentence BLEU~\cite{papineni-etal-2002-bleu} score as a proxy for n-gram overlap and compute BLEU-4 to quantify the semantic complexity of a given utterance. Figure~\ref{fig:bleu_breakdown_lm} shows that SA performance generally seems to improve for all models as lexical overlap with training utterance increases. The two-pass model is observed to be better than the baseline SLU model in all categories of semantic difficulty. Further, the performance gap between the baseline SLU model and the two-pass SLU model with DeBERTa is greatest for utterances with BLEU scores in the bucket of ($0.15$,$0.4$]. Both models seem to be similarly struggling on more challenging utterances (i.e., for BLEU$<=0.15$). Compared with the model pre-trained on the SWBD sentiment dataset, the two-pass SLU system seems particularly helpful for semantically more complex utterances.

Another way to quantify semantic difficulty is using the length of the ASR transcript for a given utterance. Figure~\ref{fig:sen_len_breakdown_lm} shows no clear trend between SA performance and transcript length for all models. However, we observe that the performance gap between the model pre-trained on the SWBD sentiment dataset and the two-pass SLU model with DeBERTa increases with an increase in transcript length. We conclude that most of the performance difference between \textit{self-supervised} LM and \textit{supervised} model trained on an external SLU dataset is for semantically challenging utterances which have low n-gram overlap with training utterances or longer utterance length.

\begin{figure}[t]
\centering
\begin{subfigure}[b]{\linewidth}
\centering
\includegraphics[width=0.9\textwidth]{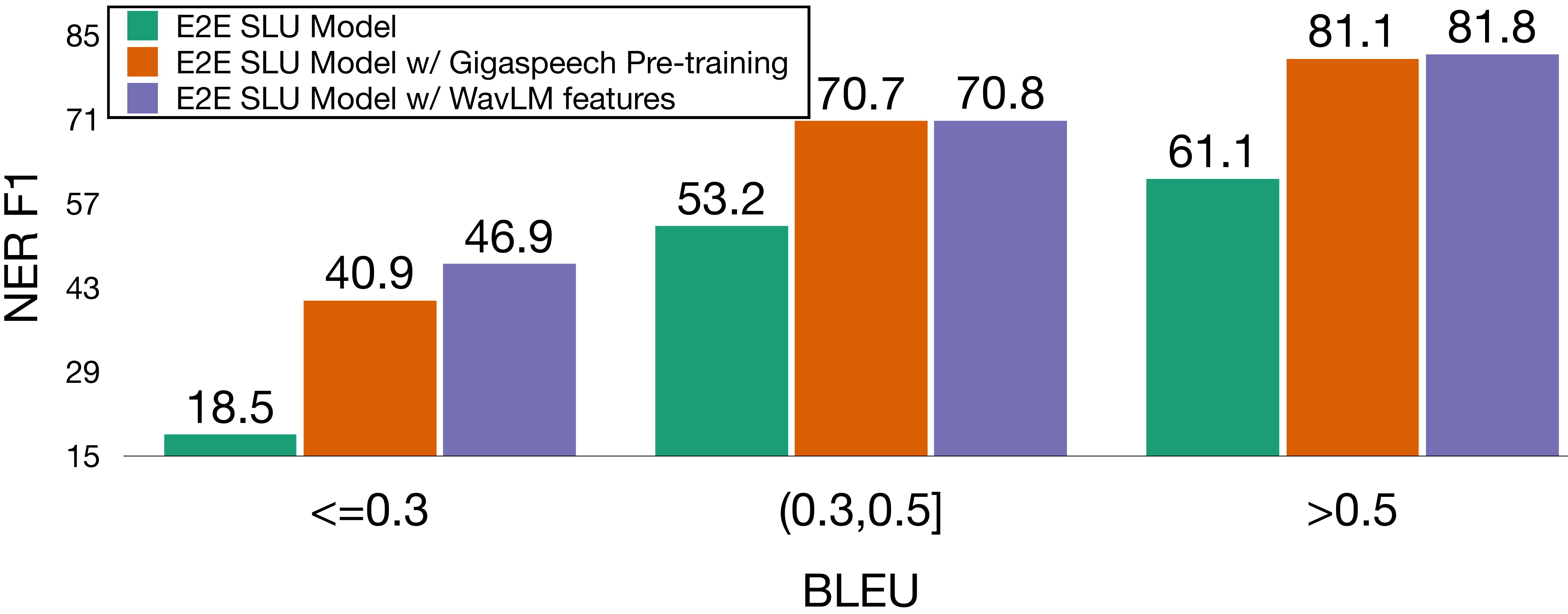}
\caption{NER performance across different lexical overlap (measured by BLEU score) of test utterances with the training set.}
\label{fig:bleu_breakdown_asr}
\end{subfigure}
\begin{subfigure}[b]{\linewidth}
\centering
\includegraphics[width=0.9\linewidth]{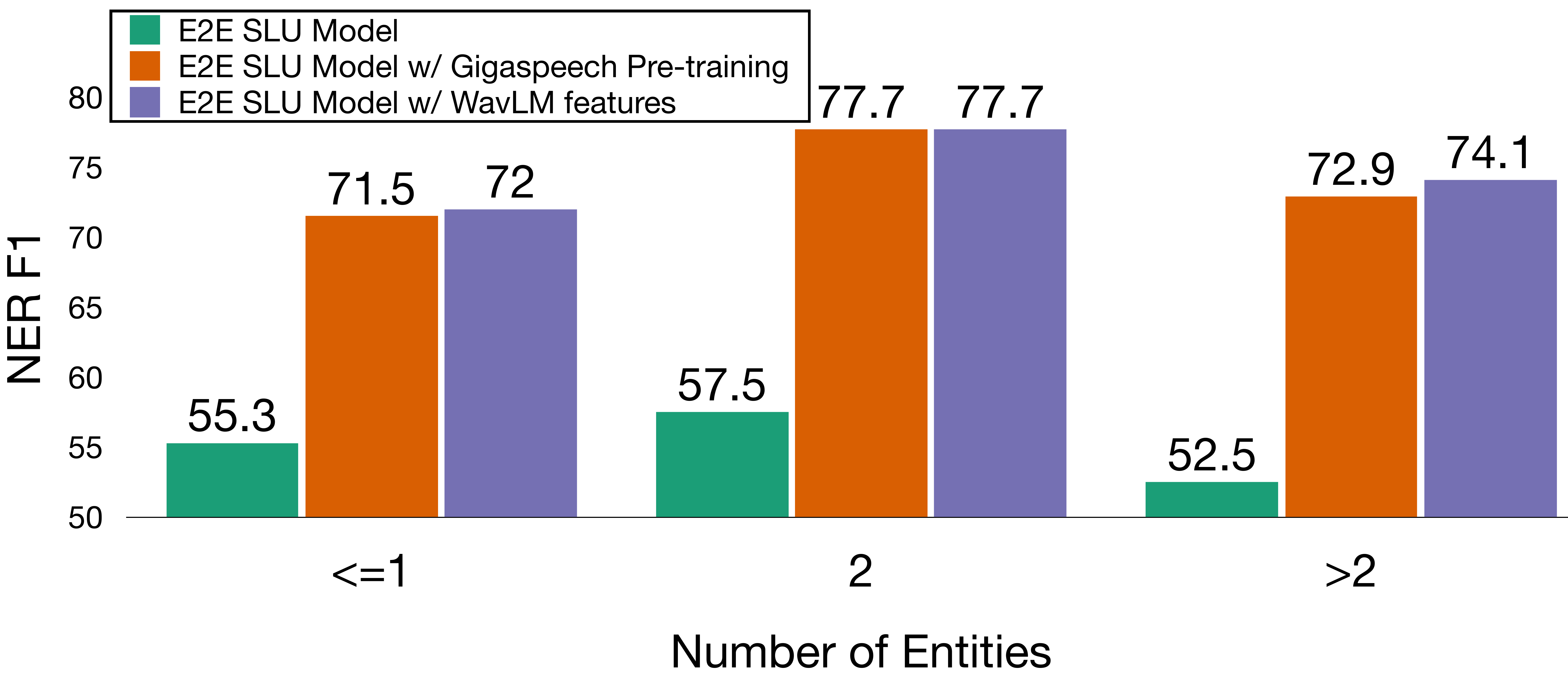}
\caption{NER performance across different number of entities in test utterances.}
\label{fig:num_ent_breakdown_asr}
\end{subfigure}
\caption{Results comparing the NER performance of models without pre-training, with ASR pre-training on GigaSpeech and with WavLM features as input across different semantic difficulties on SLUE-VoxPopuli. The semantic difficulty is measured using (a) lexical overlap with training utterances and (b) number of entities. We observe that \textit{self-supervised} WavLM representations are particularly useful for semantically complex utterances which have low n-gram overlap with the training set or many entity mentions.}
\label{fig:semantic-speech-breakdown}
\vskip -0.2in
\end{figure}

We perform a similar analysis for the NER model using WavLM features as shown in Figure~\ref{fig:semantic-speech-breakdown}. Figure~\ref{fig:bleu_breakdown_asr} shows that NER performance for all models improves with an increase in lexical overlap with training utterances. The performance gains achieved by both ASR pre-training and SSL features are highest for challenging utterances (i.e., BLEU $<=0.3$). Remarkably, most of the performance gap between WavLM and GigaSpeech model is also on utterances with high semantic complexity, probably due to extensive linguistic and acoustic variations in large amounts of unlabelled data used to train SSL speech models. Figure~\ref{fig:num_ent_breakdown_asr} breakdowns NER performance based on the number of entities in a test utterance. An utterance with many entity mentions requires better semantic understanding. Again, we observe that \textit{self-supervised} speech representations are more robust to semantic complexity (Entity No. $>2$) than \textit{supervised} representations pre-trained on an ASR dataset.

\section{Conclusion}
In this work, we present a thorough analysis of four types of pre-training approaches for SLU. We show that each of the pre-trained models can boost performance over the baseline SLU model without pre-training. Our results show that \textit{self-supervised} pre-trained models achieve higher performance than \textit{supervised} pre-trained models. Specifically, we demonstrate that SSL speech models give the most performance gains for the NER task and pre-trained LMs give the greatest performance improvement on the SA task. We also observe that gains achieved by different pre-training methodologies are complementary to each other and by combining different approaches, we can further advance the SLU performance. Finally, we show a detailed analysis to gain insights into our performance gains and infer that \textit{self-supervised} pre-trained models are particularly beneficial for acoustically and semantically challenging utterances. 

We recommend future studies to leverage \textit{self-supervised} representations to advance SLU performance, particularly for under-resourced settings. We hope insights derived from our study will facilitate future research on the tight integration of pre-training methodologies for SLU.

\section{ACKNOWLEDGMENTS}
This work used the Extreme Science and Engineering Discovery Environment (XSEDE) ~\cite{xsede}, which is supported by National Science Foundation grant number ACI-1548562. Specifically, it used the Bridges system ~\cite{nystrom2015bridges}, which is supported by NSF award number ACI-1445606, at the Pittsburgh Supercomputing Center (PSC).

\bibliographystyle{IEEEbib}
\bibliography{strings,refs}

\end{document}